\documentclass[letterpaper, 10 pt, conference]{ieeeconf}  % Comment this line out if you need a4paper

\IEEEoverridecommandlockouts                              % This command is only needed if 
\usepackage{amsmath}  % assumes amsmath package installed
\usepackage{amssymb}  % assumes amsmath package installed
\usepackage{graphicx} % for pdf, bitmapped graphics files
\usepackage{cite}  % for bibliographic citations
\usepackage{tabularx}  % for tables with fixed-width columns
\usepackage{adjustbox}  % To adjust boxes, such as scaling tables and figures
\usepackage{xcolor}  % For color text and graphics
\usepackage{makecell}  % For advanced table formatting, like multi-line cells

\title{\LARGE \bf
Traffic-Aware Pedestrian Intention Prediction
}

\author{Fahimeh Orvati Nia and Hai Lin
\thanks{This work was supported by the National Science Foundation under  Grant CNS-1830335 and Grant IIS-2007949. Fahimeh Orvati Nia and Hai Lin are both with the Department of Electrical Engineering, University of Notre Dame, Notre Dame, IN, USA. e-mail: \{{\tt forvatin, hlin1}\}{\tt @nd.edu}.}
}

\begin{document}

\maketitle

\maketitle
\vspace{-10pt}
\noindent\footnotesize\textit{This paper has been accepted to the American Control Conference (ACC) 2025.}
\vspace{10pt}

%%%%%%%%%%%%%%%%%%%%%%%%%%%%%%%%%%%%%%%%%%%%%%%%%%%%%%%%%%%%%%%%%%%%%%%%%%%%%%%%
\begin{abstract}
Accurate pedestrian intention estimation is crucial for the safe navigation of autonomous vehicles (AVs) and hence attracts a lot of research attention. However, current models often fail to adequately consider dynamic traffic signals and contextual scene information, which are critical for real-world applications. This paper presents a Traffic-Aware Spatio-Temporal Graph Convolutional Network (TA-STGCN) that integrates traffic signs and their states (Red, Yellow, Green) into pedestrian intention prediction. Our approach introduces the integration of dynamic traffic signal states and bounding box size as key features, allowing the model to capture both spatial and temporal dependencies in complex urban environments. The model surpasses existing methods in accuracy. Specifically, TA-STGCN achieves a 4.75\% higher accuracy compared to the baseline model on the PIE dataset, demonstrating its effectiveness in improving pedestrian intention prediction.
\end{abstract}

%%%%%%%%%%%%%%%%%%%%%%%%%%%%%%%%%%%%%%%%%%%%%%%%%%%%%%%%%%%%%%%%%%%%%%%%%%%%%%%%
\section{Introduction}

Pedestrians account for 23\% of global road traffic fatalities \cite{who2018}. Correctly predicting pedestrian intention is, therefore, critical for reducing accidents and enhancing both driver assistance systems and autonomous driving technologies.

Early pedestrian intention estimation methods relied on handcrafted features and classical machine learning, focusing on pedestrian position, velocity, and head orientation \cite{schulz2015}. However, these methods required extensive feature engineering and often failed to generalize across different environments \cite{keller2013}. Moreover, it is difficult to incorporate dynamic contextual information such as traffic signals and surrounding vehicles, essential for accurate predictions \cite{ghori2018}.

Deep learning has transformed pedestrian intention estimation by enabling automatic feature extraction from raw data. Convolutional Neural Networks (CNNs) and Recurrent Neural Networks (RNNs), particularly Long Short-Term Memory (LSTM) networks, have been explored for capturing spatial and temporal dependencies in pedestrian behavior \cite{rasouli2019pie}. Despite advancements, many deep learning models have been criticized for not adequately considering the broader scene context, crucial for accurate predictions in dynamic environments \cite{saleh2019}. Therefore, integrating scene context, including traffic signals and other road elements, becomes vital for advancing pedestrian intention models.

Recent studies emphasize integrating scene context into pedestrian intention models. Rasouli et al. \cite{rasouli2019pie} introduced a model incorporating local contextual information, while Cao et al. \cite{cao2020} developed a Spatio-Temporal Graph Convolutional Network (STGCN) using pedestrian skeletal information.
\begin{figure}[t]
    \centering
    \includegraphics[width=0.45\textwidth]{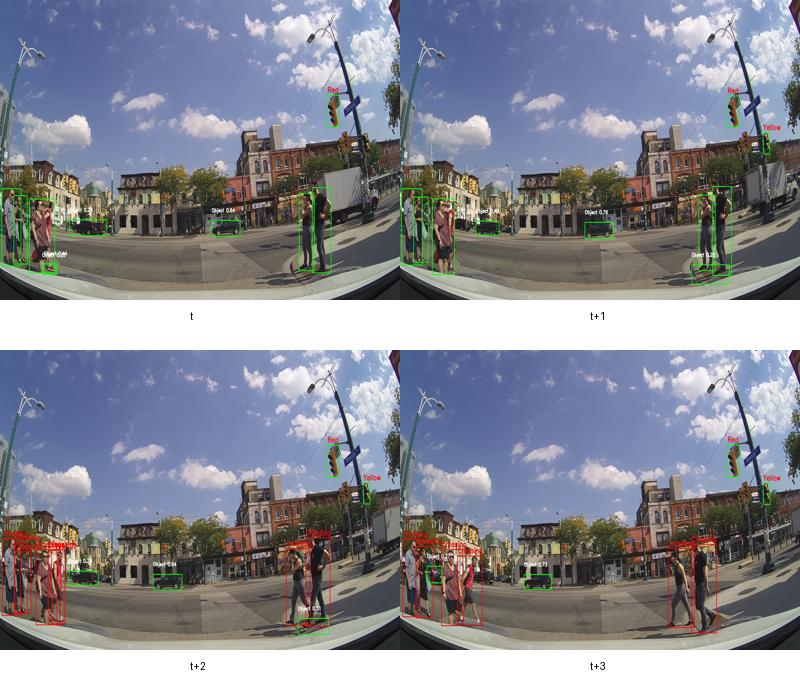}
    \caption{Pedestrian Intention Prediction Example at various time frames. In the third frame (t+2), the bounding box for the pedestrian turns red, indicating the model's prediction that the pedestrian intends to cross the street. The ground truth of this action is confirmed in the subsequent frame (t+3), where the pedestrian is indeed crossing. The system integrates bounding boxes and traffic light states (Red, Yellow) for accurate intention prediction.}
    \label{fig:pedestrian_prediction} 
\end{figure}

Prior to the development of deep learning methods, classical approaches such as the Extended Kalman Filter (EKF) were adapted for pedestrian intention prediction, estimating pedestrian states over time using motion models and sensor data \cite{schulz2015}. This approach is valuable in AVs and Advanced Driver-Assistance Systems (ADAS) but may struggle with highly nonlinear behaviors \cite{keller2013}.
Building on these advancements, we propose the TA-STGCN, a model designed to address the limitations of previous approaches by integrating dynamic traffic sign awareness and scene context. One key motivation for our work is that pedestrians' intentions are heavily influenced by the surrounding environment, particularly the states of traffic lights and crosswalks. Existing pedestrian intention models often neglect these contextual factors, which can lead to inaccurate predictions, especially in urban settings where traffic signals significantly influence pedestrian behavior.

Our model, TA-STGCN, incorporates both spatial and temporal information through the use of a Spatio-Temporal Graph Convolutional Network (STGCN). The spatial aspect is captured by modeling interactions between pedestrians and nearby objects, such as vehicles and traffic signs, while the temporal aspect considers the changes in these relationships over time. By integrating traffic light states (R, Y, G) and the spatial configuration of crosswalks into the model, we provide a more robust representation of the scene.

TA-STGCN enhances pedestrian intention prediction by explicitly incorporating traffic signals and their states into the decision-making process. This approach ensures that the model can accurately predict pedestrian behavior in complex, dynamic environments where such signals play a significant role. As a result, our model improves both the accuracy and reliability of pedestrian intention estimation, surpassing existing methods that do not fully consider these critical environmental factors.

The rest of the paper is organized as follows: Section II reviews related work. Section III defines the problem and objectives. Section IV details the methodology. Section V presents the results and a component study. Section VI discusses the model’s strengths and limitations. Section VII concludes with a summary of our findings.

%%%%%%%%%%%%%%%%%%%%%%%%%%%%%%%%%%%%%%%%%%%%%%%%%%%%%%%%%%%%%%%%%%%%%%%%%%%%%%%%
\section{Related Work}

\subsection{Traditional Approaches}

Traditional pedestrian intention estimation relied on handcrafted features and classical machine learning techniques like SVMs and decision trees \cite{keller2013}. These methods focused on features like position and velocity but required extensive feature engineering and lacked generalization across scenarios \cite{ghori2018}. It is also difficult to incorporate dynamic contextual information, which is crucial for accurate predictions.

\subsection{Extended Kalman Filter Approaches}

The Extended Kalman Filter (EKF) estimates pedestrian states over time, such as position, velocity, and intent \cite{schulz2015}. EKF fuses data from multiple sensors, making it useful in AVs for real-time pedestrian tracking. However, EKF may be less effective in scenarios with nonlinear behaviors \cite{keller2013}.

\subsection{Data-Driven Approaches}

Deep learning has significantly advanced pedestrian intention estimation. CNNs and RNNs, including LSTM networks, automatically extract and model features from raw data \cite{cao2020}. GCNs further improve by modeling complex relationships within a scene, making them suitable for this task \cite{rasouli2019pie}. Recent research underscores the need for integrating scene context, as shown by models using skeletal data \cite{cao2020}.

\subsection{Generative AI Approaches}

Generative AI methods, such as Generative Adversarial Networks (GANs) and Variational Autoencoders (VAEs), are used to model uncertainties in behavior by generating possible outcomes that account for randomness and variability \cite{sadeghian2019}. Social GAN, introduced by Gupta et al. \cite{gupta2018}, predicts realistic and socially acceptable pedestrian trajectories by considering pedestrian interactions. Sadeghian et al. \cite{sadeghian2019} extended this with Sophie, integrating social and physical constraints into pedestrian trajectory prediction. However, the model's reliance on ground truth annotations for object detection and classification limits its real-time applicability.

\subsection{Graph-Based Approaches}

Graph Convolutional Networks (GCNs) and Spatio-Temporal Graph Convolutional Networks (STGCNs) have been widely adopted for modeling interactions between objects in pedestrian intention prediction tasks. Naik et al. \cite{naik2021} proposed a Scene Spatio-Temporal Graph Convolutional Network (Scene-STGCN) that models the interactions between pedestrians and surrounding traffic objects, such as crosswalks, traffic signs, and vehicles. Their model dynamically learns the contribution levels of these surrounding objects to predict pedestrian intention.

While their approach focuses on traffic objects and general scene understanding, our proposed model (TA-STGCN) differs by explicitly incorporating traffic signal states (Red, Yellow, Green) and bounding box size as key features. These features provide a more focused approach to understanding pedestrian intention in the context of traffic signals and proximity to the autonomous vehicle.

%%%%%%%%%%%%%%%%%%%%%%%%%%%%%%%%%%%%%%%%%%%%%%%%%%%%%%%%%%%%%%%%%%%%%%%%%%%%%%%%
\section{Problem Formulation}

\subsection{Problem Definition}

This work aims to enhance pedestrian intention estimation by incorporating traffic sign awareness into an STGCN. The goal is to predict pedestrian crossing intentions and future trajectories using dynamic traffic sign information. Given a sequence of video frames, our goal is to classify pedestrian crossing intention (\( y \in \{0,1\} \)) using spatial, temporal, and contextual traffic signal features. For each video frame, three primary sources of information are extracted:

\begin{itemize}
    \item \textbf{Image Data \( I \):} Contains images of \( N \) objects per frame (\( H \times W \)) over \( T \) time frames.
    \item \textbf{Location Data \( L_f \):} Stores 2D coordinates of object bounding boxes over \( T \) frames.
    \item \textbf{Class Data \( C_f \):} Includes object class labels and traffic signal states (R, Y, G) as one-hot vectors over \( T \) frames.
\end{itemize}

These features are extracted using ground truth annotations to ensure that the model focuses on intention estimation performance without being affected by potential errors in object detection and classification. The availability of annotated data plays a crucial role in the effectiveness of such models. In this case, datasets like PIE provide high-quality annotations for pedestrians and traffic signals, allowing for accurate and reliable intention prediction. However, in real-world scenarios where annotated data may not always be available, the performance of the model could degrade due to potential errors in object detection and classification. In such cases, alternative strategies like weak supervision, transfer learning, or the use of semi-supervised techniques could help bridge the gap, enabling the model to still make reasonable predictions despite the absence of fully annotated data. Thus, while ground truth annotations provide the ideal setup for model training, future work could explore more robust methods to handle scenarios where such data is limited or unavailable.

\subsection{Model Objective}

The proposed TA-STGCN estimates the crossing intention \( \hat{y} \) using inputs \( I \) (image data), \( L_f \) (location features), and \( C_f \) (class labels):
\vspace{-5pt}
\[
\hat{y} = \text{TA-STGCN}(I, L_f, C_f)
\]

The model is optimized by minimizing the binary cross-entropy loss function, which is widely used for classification tasks involving two classes (e.g., crossing vs. not crossing). The binary cross-entropy loss function is defined as:

\[
L(\hat{y}, y) = -\left[ y \log(\hat{y}) + (1 - y) \log(1 - \hat{y}) \right]
\]

where \( y \in \{0, 1\} \) represents the true intention, and \( \hat{y} \) is the predicted probability of the pedestrian intending to cross. The objective is to minimize this loss function, which penalizes large deviations between the predicted probability and the true label, ensuring that the model learns to predict crossing intentions as accurately as possible.

%%%%%%%%%%%%%%%%%%%%%%%%%%%%%%%%%%%%%%%%%%%%%%%%%%%%%%%%%%%%%%%%%%%%%%%%%%%%%%%%
\section{Methodology}

In this section, we provide an in-depth explanation of the proposed model architecture and the reasoning behind each component choice. The goal is to justify why specific methods were selected and how they contribute to the overall performance of pedestrian intention prediction.

\subsection{System Overview}

\begin{figure*}[h]
    \centering
    \includegraphics[width=0.75\linewidth, , height = 4cm]{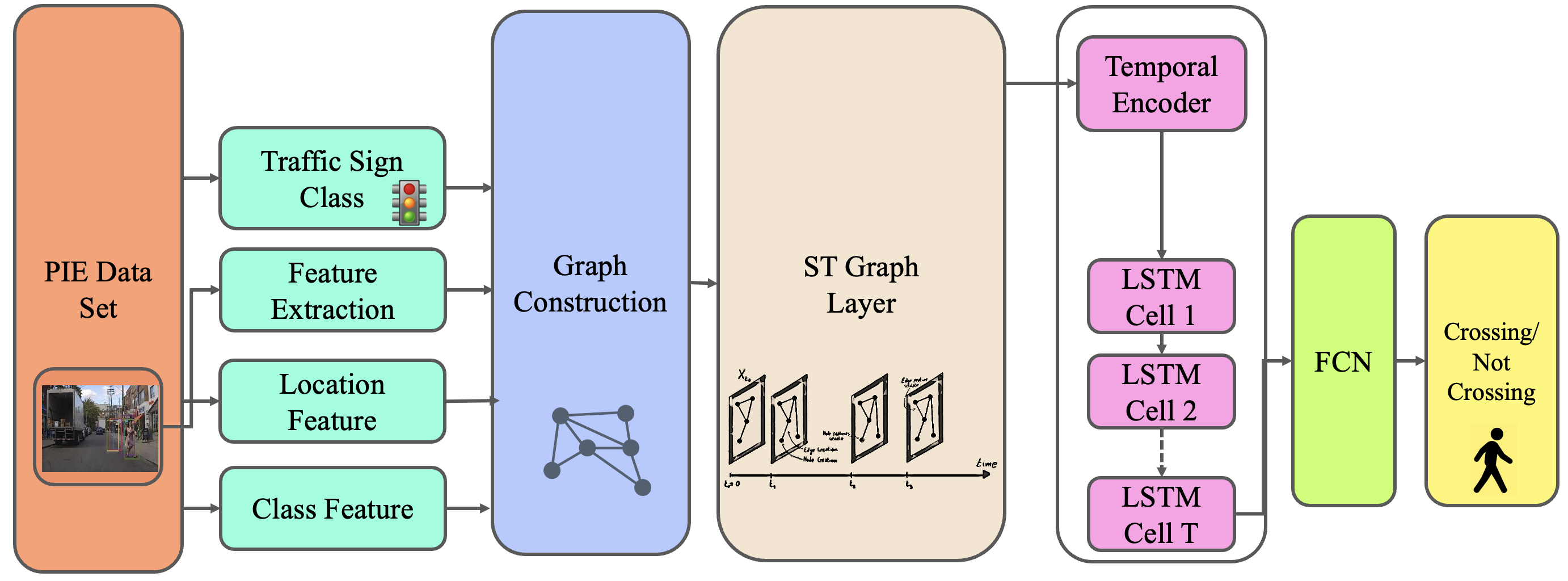}
    \caption{Overview of the proposed TA-STGCN architecture. The model processes two feature streams: the image-class stream (combining image and class feature data) and the location-class stream (combining location and class feature data). Each stream is separately processed by ST-Graph layers to encode spatial and short-term temporal dependencies. The outputs are then concatenated and passed into a temporal encoder with LSTM cells to capture long-term dependencies, leading to a crossing or not-crossing decision through a Fully Connected Network (FCN).}
    \label{fig:model}
\end{figure*}

The overview of the proposed TA-STGCN model is illustrated in Figure \ref{fig:model}. The model processes two streams of features that jointly capture both spatial and temporal aspects of the scene. This design choice allows the model to encode complex interactions between pedestrians and dynamic elements in the environment, such as traffic lights, which are crucial for accurately predicting crossing intentions.

The rationale behind choosing this architecture is threefold:
\begin{itemize}
    \item \textbf{Multi-Stream Processing}: By separating image-class and location-class streams, we ensure that spatial (location) and semantic (class) information are treated differently, thereby allowing the model to capture both the physical movement and semantic properties of objects in the scene.
    \item \textbf{Spatio-Temporal Graph Layers}: These layers allow us to model complex interactions between nodes (pedestrians, traffic lights) across both space and time. This is critical for pedestrian intention prediction in urban environments, where interactions between pedestrians and surrounding traffic elements are highly dynamic.
    \item \textbf{LSTM Temporal Encoder}: Long Short-Term Memory (LSTM) units are selected to capture long-term dependencies, which are essential for correctly predicting pedestrian actions based on past motion and contextual information. LSTMs are chosen for their ability to handle temporal dependencies better than simpler RNNs.
\end{itemize}

\subsection{Intention and Trajectory Prediction}

The TA-STGCN model outputs both the crossing probability \( \hat{y} \) and the future pedestrian position \( \hat{\mathbf{p}}_{t+\Delta t} \). The sigmoid function \( \sigma \) is used for crossing probability estimation, while a fully connected output layer predicts future positions:
\vspace{-6pt}
\begin{eqnarray}
    \hat{y} &=& \sigma(\mathbf{w}^\top \mathbf{h}_v), \\
    \hat{\mathbf{p}}_{t+\Delta t} &=& \mathbf{W}_o \mathbf{h}_v
\end{eqnarray}
\vspace{-5pt}
Where:
\begin{itemize}
    \item \( \hat{y} \) is the predicted crossing probability.
    \item \( \mathbf{h}_v \) is the hidden pedestrian representation.
    \item \( \mathbf{w}^\top \) and \( \mathbf{W}_o \) are the weight vectors for crossing prediction and future position estimation, respectively.
\end{itemize}

The inclusion of trajectory prediction ensures that our model is capable of predicting not only whether a pedestrian will cross but also where they will likely move, providing a more comprehensive understanding of pedestrian behavior.

\subsection{Graph Construction}

We model the interactions between pedestrians and traffic elements (e.g., traffic lights, vehicles) using a spatio-temporal graph \( \mathcal{G} = (\mathcal{V}, \mathcal{E}) \), where the nodes represent pedestrians and traffic elements, and the edges represent interactions over time. The graph is defined as follows:
\vspace{-5pt}
\[
\mathcal{V} = \{p_t^i | t \in [1, T], i \in [1, N]\} \cup \{s_t^j | t \in [1, T], j \in [1, M]\}
\]

The adjacency matrix \( A \), representing connections between nodes, is normalized to stabilize the training process. We also introduce a parametric matrix \( W \) to learn the relative importance of different interactions. This allows the model to focus on key interactions, such as proximity to traffic lights or the presence of vehicles.

The choice of a spatio-temporal graph is motivated by the need to model interactions in both space and time. This enables the model to better capture how pedestrian behavior changes based on the environment and traffic dynamics.

\subsection{Feature Representation}

Each node in the graph is assigned a feature vector that represents either a pedestrian or a traffic sign. The feature vectors are defined as follows:
\vspace{-5pt}
\begin{eqnarray}
    \mathbf{f}_p &=& \text{CNN}(I_t, p_t^i), \\
    \mathbf{f}_s &=& [\text{type}(s_t^j), \text{state}(s_t^j), \text{class}(s_t^j), \text{pos}(s_t^j)]
\end{eqnarray}

Where:
\begin{itemize}
    \item \( \mathbf{f}_p \)Pedestrian features (position, velocity, appearance, bounding box size) extracted via CNN.
    \item \( \mathbf{f}_s \)Traffic sign features (type, state, class, position).
\end{itemize}

We include bounding box size as a feature to capture pedestrian proximity to the vehicle. As the bounding box increases in size, it indicates that the pedestrian is approaching the vehicle. This feature is crucial for detecting whether the pedestrian intends to cross and helps the model make more accurate predictions by leveraging both spatial and temporal information.

\subsection{Modeling Dependencies}

To model dependencies between nodes, we use Graph Convolutional Networks (GCNs) for spatial dependencies and Temporal Convolutional Networks (TCNs) for temporal dependencies. The updates for each node \( v \) at layer \( l+1 \) are given by:
\vspace{-5pt}
\[
\mathbf{h}_v^{(l+1)} = \sigma \left( \sum_{u \in \mathcal{N}(v)} \frac{1}{\sqrt{d_v d_u}} \mathbf{W}^{(l)} \mathbf{h}_u^{(l)} \right)
\]
\vspace{-5pt}
Where:
\begin{itemize}
    \item \( \mathbf{h}_v^{(l+1)} \) is the updated feature vector for node \( v \).
    \item \( \mathcal{N}(v) \) represents the neighboring nodes of \( v \).
    \item \( d_v \) and \( d_u \) are the degrees of nodes \( v \) and \( u \), used to normalize the contributions from neighboring nodes.
    \item \( \mathbf{W}^{(l)} \) is the weight matrix for layer \( l \).
\end{itemize}

Temporal dependencies are modeled using 1D convolutions along the time axis. The updated feature vector for node \( v \) at time \( t+1 \) is computed as:
\vspace{-5pt}
\[
\mathbf{h}_v^{(t+1)} = \text{ReLU} \left( \sum_{\tau=0}^{k-1} \mathbf{W}_\tau \mathbf{h}_v^{(t-\tau)} \right)
\]

Where \( k \) is the kernel size for the convolution. This design allows the model to efficiently capture both spatial and temporal relationships in the data.

\subsection{Spatio-Temporal Graph Layer}

The Spatio-Temporal Graph (ST-Graph) layer consists of three modules:

\begin{enumerate}
    \item \textbf{Feature Convolution (FC)}: Extracts deeper feature representations using \( 1 \times 1 \) convolutions, which preserve spatial and temporal dimensions.
    \item \textbf{Spatial Message Passing (SMP)}: Updates node features based on neighboring nodes, using the adjacency matrix \( A \). An identity matrix \( I_N \) is added for self-connections, and the normalized matrix \( \tilde{A}_N = D^{-1/2} \tilde{A} D^{-1/2} \) ensures stable updates.
    \item \textbf{Temporal Message Passing (TMP)}: Captures short-term temporal dependencies using 1D convolutions, ensuring that the temporal dimension is preserved.
\end{enumerate}

\subsection{Temporal Encoder}

The concatenated outputs from the ST-Graph layers are processed by the temporal encoder, which consists of a sequence of Long Short-Term Memory (LSTM) cells. This design is motivated by the need to capture long-term temporal dependencies, which are essential for predicting future pedestrian behavior. The final output of the LSTM cells is passed into a Fully Connected Network (FCN) to make the final crossing intention prediction.

%%%%%%%%%%%%%%%%%%%%%%%%%%%%%%%%%%%%%%%%%%%%%%%%%%%%%%%%%%%%%%%%%%%%%%%%%%%%%%%%
\section{Results and Evaluation}

\subsection{Dataset}

The model is evaluated on the PIE dataset \cite{rasouli2019pie}, which includes over 6 hours of annotated driving videos captured in urban environments at 30 FPS with a resolution of 1920x1080 pixels. The dataset provides annotations for traffic elements like pedestrians, vehicles, and crosswalks. The observation length is 15 frames (0.5 seconds), and the dataset is split into training, validation, and test sets using standard splits.

\begin{figure}[h]
    \centering
    \includegraphics[width=0.7\linewidth]{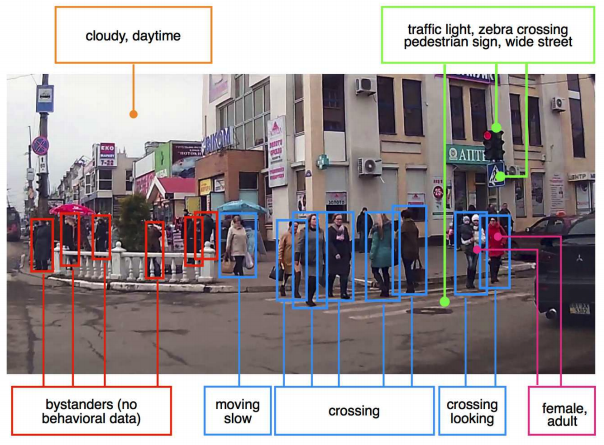}
    \caption{Example images from the PIE dataset.}
    \label{fig:pie_dataset}
\end{figure}

\subsection{Training Process}

The model was trained on the PIE dataset using standard training, validation, and test splits. The dataset consists of sequences of 15 frames (0.5 seconds) captured at 30 FPS. The data includes detailed annotations to isolate pedestrian intention estimation from object detection, allowing a focused evaluation of the TA-STGCN model.

The TA-STGCN model was trained using binary cross-entropy loss, which is suitable for predicting pedestrian crossing intention. The model was optimized with stochastic gradient descent (SGD) with learning rates of 5e-4, 7e-4, and 1e-3, depending on the number of nodes. L1 and L2 regularization were applied to prevent overfitting: L1 with a coefficient of 0.01 for LSTM cells and L2 with coefficients of 0.05 and 0.001 for STGCN layers in the image-class and location-class streams, respectively.
The model was trained with a batch size of 128 on NVIDIA Tesla V100 GPUs. The training process was repeated five times for robustness, and mean results were reported. The evaluation metrics used are detailed in Table \ref{tab:evaluation_metrics}.
\vspace{-5pt}
\begin{table}[h]
\centering
\caption{Evaluation Metrics}
\begin{tabular}{|>{\raggedright}p{0.11\linewidth}|>{\raggedright}p{0.15\linewidth}|>{\raggedright\arraybackslash}p{0.45\linewidth }|}
\hline
\textbf{Metric} & \textbf{Description} & \textbf{Formula} \\ \hline
Accuracy & Overall correctness & \[\frac{\text{TP} + \text{TN}}{\text{Total Instances}}\] \\ \hline
Precision &  Accuracy of positive predictions & \[\frac{\text{TP}}{\text{TP} + \text{FP}}\] \\ \hline
Recall & Ability to find relevant cases & \[\frac{\text{TP}}{\text{TP} + \text{FN}}\] \\ \hline
F1-Score & Balances Precision and Recall & \[\text{2} \times \frac{\text{Precision} \times \text{Recall}}{\text{Precision} + \text{Recall}}\] \\ \hline
ADE & Average prediction error & \[\frac{1}{T} \sum_{t=1}^{T} \sqrt{(x_t - \hat{x}_t)^2 + (y_t - \hat{y}_t)^2}\] \\ \hline
FDE & Final prediction accuracy & \[\sqrt{(x_T - \hat{x}_T)^2 + (y_T - \hat{y}_T)^2}\] \\ \hline
\end{tabular}
\label{tab:evaluation_metrics}
\end{table}
\vspace{-5pt}
\subsection{Performance Evaluation}

Table \ref{table:performance_comparison} summarizes the performance of different models on the PIE dataset using the proposed evaluation metrics.

\begin{table*}[htbp]
\caption{Performance on the PIE Dataset}
\centering
\begin{tabularx}{1\linewidth}{|X|c|c|c|c|c|c|}
\hline
\textbf{Model} & \textbf{Accuracy} & \textbf{Precision} & \textbf{Recall} & \textbf{F1-Score} & \textbf{ADE} & \textbf{FDE} \\
\hline
SVM & 65.34\% & 63.58\% & 61.74\% & 62.65\% & 0.95 & 1.97 \\
RNN & 76.89\% & 74.50\% & 72.87\% & 73.68\% & 0.72 & 1.39 \\
GCN & 79.00\% & 77.50\% & 76.12\% & 76.80\% & 0.58 & 1.15 \\
PIE Baseline Model & 79.00\% & 84.00\% & 85.00\% & 84.50\% & 0.47 & 0.92 \\
\textbf{Ours} & \textbf{84.65\%} & \textbf{86.50\%} & \textbf{88.03\%} & \textbf{87.26\%} & \textbf{0.43} & \textbf{0.91} \\
\hline
\end{tabularx}
\label{table:performance_comparison}
\end{table*}

\subsection{Analysis of Results}

As shown in Table \ref{table:performance_comparison}, TA-STGCN outperforms traditional methods like SVM and RNN, showing a marked improvement in both accuracy and trajectory prediction metrics. While the GCN model performs better than SVM and RNN, our TA-STGCN model achieves the highest overall performance, underscoring the effectiveness of integrating traffic sign awareness into pedestrian intention prediction.

\subsection{Component Analysis}

A comprehensive component analysis was conducted to assess the impact of integrating traffic sign information, specifically their classification (e.g., green, red, yellow), into the Temporal Attention-based Spatio-Temporal Graph Convolutional Network (TA-STGCN) model. The objective was to determine whether the inclusion of this dynamic contextual information could enhance the model's predictive performance.

Table \ref{table:ablation_study} presents the results of this analysis, comparing the original STGCN model with the modified TA-STGCN that incorporates traffic sign data. The metrics evaluated include Accuracy, Precision, Recall, F1-Score, Average Displacement Error (ADE), and Final Displacement Error (FDE).

\begin{table}[htbp]
\caption{Component Analysis on the Effect of Traffic Sign Integration}
\centering
\begin{tabularx}{1\linewidth}{|>{\raggedright\arraybackslash}X|>{\centering\arraybackslash}c|>{\centering\arraybackslash}c|>{\centering\arraybackslash}c|>{\centering\arraybackslash}c|>{\centering\arraybackslash}c|>{\centering\arraybackslash}c|}
\hline
\textbf{Model} & \makecell{\textbf{Accu-} \\ \textbf{racy}} & \makecell{\textbf{Preci-} \\ \textbf{sion}} & \makecell{\textbf{Re-} \\ \textbf{call}} & \makecell{\textbf{F1-} \\ \textbf{Score}} & \makecell{\textbf{ADE}} & \makecell{\textbf{FDE}} \\
\hline
STGCN & 83.30\% & 84.20\% & 86.00\% & 85.20\% & 0.44 & 0.93 \\
\textbf{TA-STGCN} & \textbf{84.65\%} & \textbf{86.50\%} & \textbf{88.03\%} & \textbf{87.27\%} & \textbf{0.43} & \textbf{0.91}  \\
\hline
\end{tabularx}
\label{table:ablation_study}
\end{table}

The results indicate that the TA-STGCN model, with the integration of traffic sign information, outperforms the baseline STGCN model across all evaluated metrics. This improvement underscores the importance of incorporating dynamic environmental factors, such as traffic signs, in enhancing the model's ability to accurately predict pedestrian behavior. Specifically, the increases in Precision, Recall, and F1-Score suggest a more robust performance in identifying and tracking pedestrian intentions, while the reductions in ADE and FDE reflect more accurate trajectory predictions.

\subsection{Quantitative Analysis}

\begin{figure}[h]
    \centering
    \includegraphics[width=0.8\linewidth]{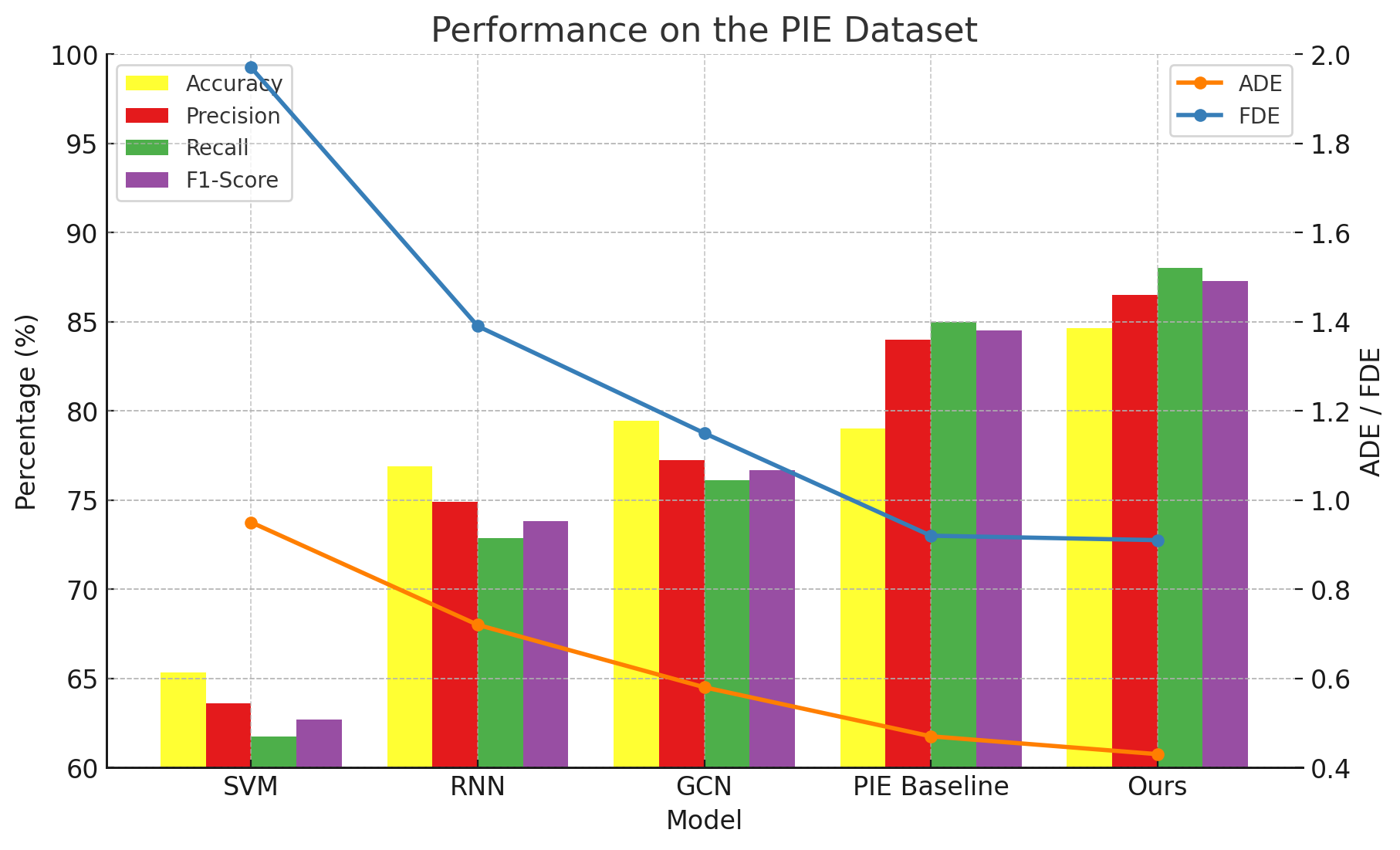}
    \caption{Performance comparison of various models on the PIE dataset across different metrics. The TA-STGCN model outperforms other models in both classification (Accuracy, Precision, Recall, F1-Score) and trajectory prediction (ADE, FDE).}
    \label{fig:quantitative_analysis}
\end{figure}

Figure \ref{fig:quantitative_analysis} presents a comprehensive quantitative analysis comparing different models on the PIE dataset across multiple evaluation metrics. The models compared include SVM, RNN, GCN, the PIE baseline model, and our proposed TA-STGCN model. Each bar in the figure represents a key metric—Accuracy, Precision, Recall, and F1-Score—while the line plots represent Average Displacement Error (ADE) and Final Displacement Error (FDE).

The left vertical axis shows the percentage values for the classification metrics (Accuracy, Precision, Recall, and F1-Score), and the right vertical axis shows the ADE/FDE values. A higher bar indicates better performance for the classification metrics, while lower values for ADE and FDE suggest more accurate trajectory predictions.

The TA-STGCN model shows significant improvements across all metrics. Specifically, it achieves the highest Accuracy (84.65\%) compared to other models, such as the PIE baseline (79.00\%) and GCN (79.00\%). The improvements in Precision (86.50\%) and Recall (88.03\%) indicate that TA-STGCN can better identify true pedestrian crossing intentions while reducing false predictions. Moreover, the F1-Score of 87.27\% demonstrates a balance between Precision and Recall, further validating the model's robustness.

Additionally, the ADE and FDE lines show that our model substantially reduces the prediction error. With an ADE of 0.43 and an FDE of 0.91, TA-STGCN outperforms the baseline model and other approaches, demonstrating its ability to predict pedestrian trajectories with higher accuracy.

These results underscore the effectiveness of integrating traffic sign awareness and spatio-temporal graph convolution in enhancing both the intention prediction and trajectory forecasting of pedestrians in urban environments.

\vspace{-5pt}
%%%%%%%%%%%%%%%%%%%%%%%%%%%%%%%%%%%%%%%%%%%%%%%%%%%%%%%%%%%%%%%%%%%%%%%%%%%%%%%%
\section{Discussion}

TA-STGCN significantly improves pedestrian intention prediction, outperforming existing methods on the PIE dataset by integrating traffic sign awareness and context.

However, the model has limitations that need further discussion. One critical aspect of real-time deployment is how to extract bounding box information, object classes, and other features from raw video frames. Currently, the model relies on pre-annotated datasets for object detection and classification. In real-world applications, this information must be generated dynamically using object detection models. The accuracy and speed of these models impact TA-STGCN real-time performance, making efficient object detection essential.

Moreover, real-time computation remains a challenge. Although the TA-STGCN model shows improvements in prediction accuracy, its spatio-temporal graph construction and LSTM layers introduce computational overhead, particularly in systems with limited resources, such as embedded devices for autonomous driving. Optimizing the model for computational efficiency without sacrificing prediction performance is crucial for deploying the system in real time.

Another limitation is the reliance on ground truth annotations for object detection and classification. While this helps isolate pedestrian intention prediction in the research context, real-time systems must cope with noisy or incomplete data. Future research could explore methods to mitigate the effects of misclassification or incomplete bounding box data on the intention prediction pipeline.

Additionally, the current dataset lacks coverage of some critical traffic scenarios, such as bus stops and shared spaces where pedestrians may behave differently. Addressing this issue requires expanding dataset diversity to better generalize the model across various traffic environments.

Future work should address these limitations by enhancing feature extraction speed for real-time deployment and expanding dataset diversity for better generalization. Leveraging lightweight neural networks or edge computing can improve efficiency, making TA-STGCN viable for real-time AV systems.
\vspace{-5pt}

%%%%%%%%%%%%%%%%%%%%%%%%%%%%%%%%%%%%%%%%%%%%%%%%%%%%%%%%%%%%%%%%%%%%%%%%%%%%%%%%
\section{Conclusion}

This paper introduces TA-STGCN, which integrates traffic sign awareness with scene context to enhance pedestrian intention prediction for autonomous vehicles. The model shows significant improvements in accuracy, precision, and other key metrics, as validated by the PIE dataset, underscoring the value of dynamic traffic sign information.

TA-STGCN enhances pedestrian intention prediction and autonomous driving safety, but future work should improve real-time application and dataset diversity.
%%%%%%%%%%%%%%%%%%%%%%%%%%%%%%%%%%%%%%%%%%%%%%%%%%%%%%%%%%%%%%%%%%%%%%%%%%%%%%%%


\begin{thebibliography}{99}

\bibitem{who2018} World Health Organization, ``Global status report on road safety 2018: summary,'' World Health Organization, 2018.

\bibitem{schulz2015} A. T. Schulz and R. Stiefelhagen, ``Pedestrian intention recognition using latent-dynamic conditional random fields,'' in \emph{Proc. IEEE Intell. Vehicles Symp.}, 2015, pp. 622--627.

\bibitem{keller2013} C. G. Keller and D. M. Gavrila, ``Will the pedestrian cross? A study on pedestrian path prediction,'' \emph{IEEE Trans. Intell. Transp. Syst.}, vol. 15, no. 2, pp. 494--506, 2013.

\bibitem{ghori2018} O. Ghori, R. Mackowiak, M. A. Bautista, N. Beuter, L. Drumond, and B. Ommer, ``Learning to forecast pedestrian intention from pose dynamics,'' in \emph{Proc. IEEE Intell. Vehicles Symp.}, 2018, pp. 1272--1279.

\bibitem{naik2021} K. Naik, S. Chandra, and M. Shehata, "Scene Spatio-Temporal Graph Convolutional Network for Pedestrian Intention Estimation," in *2021 IEEE International Conference on Robotics and Automation (ICRA)*, Xi'an, China, 2021, pp. 1324-1330.

\bibitem{rasouli2019pie} A. Rasouli, I. Kotseruba, T. Kunic, and J. K. Tsotsos, ``PIE: A large-scale dataset and models for pedestrian intention estimation and trajectory prediction,'' in \emph{Proc. IEEE/CVF Int. Conf. Comput. Vis.}, 2019, pp. 6262--6271.

\bibitem{saleh2019} K. Saleh, M. Hossny, and S. Nahavandi, ``Real-time intent prediction of pedestrians for autonomous ground vehicles via spatio-temporal densenet,'' in\emph{Proc. IEEE Int. Conf. Robot. Autom.}, 2019, pp. 970--976.

\bibitem{cao2020} D. Cao and Y. Fu, ``Using Graph Convolutional Networks Skeleton-Based Pedestrian Intention Estimation Models for Trajectory Prediction,'' in \emph{J. Phys. Conf. Ser.}, vol. 1621, no. 1, pp. 012003, 2020.

\bibitem{gupta2018} A. Gupta, J. Johnson, L. Fei-Fei, S. Savarese, and A. Alahi, ``Social GAN: Socially acceptable trajectories with generative adversarial networks,'' in \emph{Proc. IEEE Conf. Comput. Vis. Pattern Recognit.}, 2018, pp. 2255--2264.

\bibitem{sadeghian2019} A. Sadeghian, V. Kosaraju, A. Sadeghian, S. Savarese, and H. Rezatofighi, ``Sophie: An attentive GAN for predicting paths compliant to social and physical constraints,'' in \emph{Proc. IEEE Conf. Comput. Vis. Pattern Recognit.}, 2019, pp. 1349--1358.

\end{thebibliography}
\end{document}